\documentclass[journal]{IEEEtran}
\usepackage{booktabs} 
\usepackage{enumerate}
\usepackage{booktabs}
\usepackage{lettrine}

\usepackage{cite}

\ifCLASSINFOpdf
   \usepackage[pdftex]{graphicx}
   \usepackage{subfigure}
\else
\fi
\usepackage{amsmath}
\usepackage{bbm}
\usepackage{algorithmic}
\usepackage{algorithm}
\begin{document}
%
\title{Progress Regression RNN for Online Spatial-Temporal Action Localization in Unconstrained Videos}
%
%
%

\author{Bo~Hu, 
        
        Jianfei~Cai,~\IEEEmembership{Senior Member,~IEEE,}
        Tat-Jen~Cham 
        and~Junsong~Yuan,~\IEEEmembership{Senior Member,~IEEE,}
\thanks{B. Hu, J. Cai, and T. Cham are with the School of Computer Science and Engineering, Nanyang Technological University, Singapore 639798 (e-mail: hubo@ntu.edu.sg; asjfcai@ntu.edu.sg; astjcham@ntu.edu.sg)}

\thanks{J. Yuan is with the Department of Computer Science and Engineering, University of Buffalo, Buffalo, NY 14260-2500, USA. (e-mail: jsyuan@buffalo.edu)} 
}

%
%

\markboth{Journal of \LaTeX\ Class Files,~Vol.~14, No.~8, August~2015}%
{Shell \MakeLowercase{\textit{et al.}}: Bare Demo of IEEEtran.cls for IEEE Journals}
%



\maketitle

\begin{abstract}
Previous spatial-temporal action localization methods commonly follow the pipeline of object detection to estimate bounding boxes and labels of actions. However, the temporal relation of an action has not been fully explored.
In this paper, we propose an end-to-end Progress Regression Recurrent Neural Network (PR-RNN) for online spatial-temporal action localization, which learns to infer the action by temporal progress regression. Two new action attributes, called progression and progress rate, are introduced to describe the temporal engagement and relative temporal position of an action. 
In our method, frame-level features are first extracted by a Fully Convolutional Network (FCN). Subsequently, detection results and action progress attributes are regressed by the Convolutional Gated Recurrent Unit (ConvGRU) based on all the observed frames instead of a single frame or a short clip. Finally, a novel online linking method is designed to connect single-frame results to spatial-temporal tubes with the help of the estimated action progress attributes. 
Extensive experiments demonstrate that the progress attributes improve the localization accuracy by providing more precise temporal position of an action in unconstrained videos. Our proposed PR-RNN achieves the state-of-the-art performance for most of the IoU thresholds on two benchmark datasets.
\end{abstract}


%
\IEEEpeerreviewmaketitle

\section{Introduction}\label{secIntro}
\lettrine[lines=2]{A}{CTION}
Analysis is one of the most popular tasks in video analytics. 
In the past few years, most of the research efforts have concentrated on the task of action recognition \cite{wang2013action,wang2015action,tran2015learning,fernando2015modeling,carreira2017quo}, which predicts an action label for a trimmed video. However, in the real world scenarios, such as video surveillance \cite{oh2011large,hu2004survey} and human-computer interaction \cite{murphy2010human}, trimmed videos are usually not provided. Thus, when and where the target action appears are more essential for further analysis. 
Online spatial-temporal action localization aims to detect the spatial-temporal locations of actions in an ongoing video stream. In this task, several action tubes are generated for a testing video in an online manner. Each of the action tube consists of a sequence of bounding boxes which are connected across frames. It is a challenging problem due to large intra-class variation, insufficient action observations, and complicated background clutter in both spatial and temporal domain.

\begin{figure}[t]
    \centering
    \includegraphics[width=0.48\textwidth]{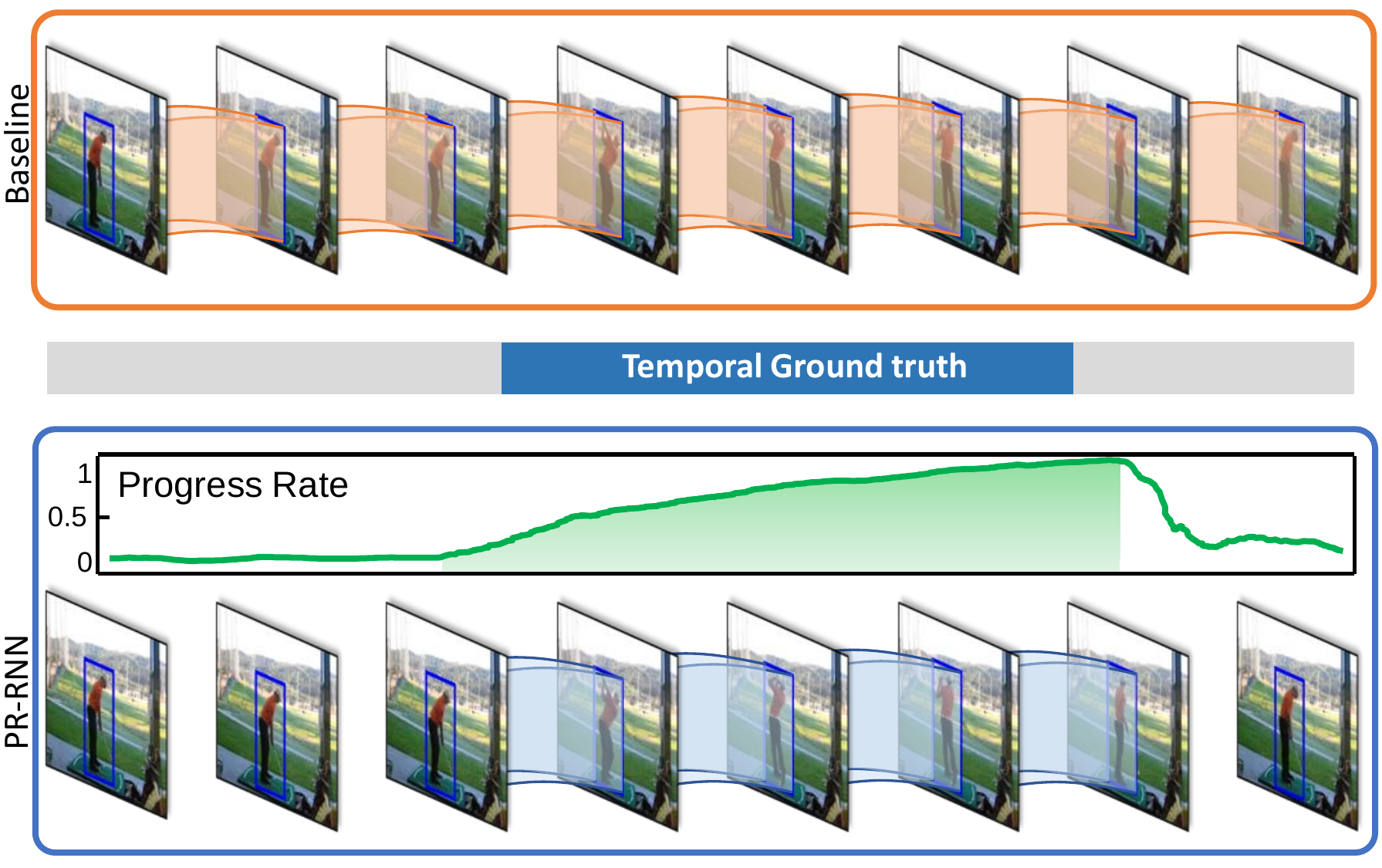}
    \caption{Illustration of the idea of PR-RNN detector. Our detector assigns multiple class-specific progress rates to each predicted bounding box. Temporal labeling in action tube generation is achieved by finding the increasing progress rate sequence (green shadow), which significantly improves the temporal localization accuracy of action tubes.}
    \label{FigOverview}
\end{figure}

With the development of Convolutional Neural Network (CNN) based object detectors \cite{girshick2015fast,liu2016ssd,redmon2017yolo9000,ren2015faster}, impressive achievements have been made in spatial-temporal action localization \cite{gkioxari2015finding,peng2016multi,hou2017tube,weinzaepfel2015learning}. The recently proposed approaches either exploit a CNN detector directly to localize action instances in every single frame \cite{peng2016multi} or improve the detector by expanding the input for the network, such as multi-frame stacking \cite{kalogeiton2017action} and clip input with 3D convolution \cite{hou2017tube}. All of these detectors produce the same type of outputs: i) an actionness score of a bounding box proposal; ii) the coordinate offsets for bounding box refinement; iii) classification probabilities for all the action categories. These methods have achieved remarkable performance, however, they did not fully explore the difference between objects and actions. 
Unlike objects in images, actions have temporal structures. Temporal relation among different frames of an action is not fully exploited in these methods. Furthermore, a single actionness score cannot accurately distinguish the action from complex background especially in temporal domain. Thus, to better locate actions in both spatial and temporal domain, an action detector should also tell the temporal progress of an action, \textit{e.g.} the action is in progress or not, just starts or is going to complete. 
Taking the ``golf swing'' action in Figure \ref{FigOverview} as an example, in the 4-th frame, the golf player is swing the club to the top, from which we can infer that this action of ``golf swing'' is in progress and has been performed about 50\%. In the first two frames, the player is just aiming the ball, however, a single actionness score usually fails to distinguish irrelevant actions from target actions, which results in not only false positive detection results, but also inaccurate temporal boundary of action tubes. Temporal progress modeling is vital to online action recognition and detection, as it can help predict what will happen next. 


In this paper, we propose an end-to-end Progress Regression Recurrent Neural Network (PR-RNN) detector. Our detector improves the previous action detectors by adding the detection of the temporal progress of actions, which is represented by two extra attributes for every frame-level action instance. The first is the progression, which indicates the probability of the target action being performed. 
It helps to eliminate the false positive detection results when a high actionness score is assigned to an irrelevant action. 
The second attribute is the progress rate, which indicates the progress proportion of the ongoing target action. During the training stage, the supervision of these two attributes allows our detector to infer the temporal status for every single frame of actions. 
The backbone network of YOLOv2 \cite{redmon2017yolo9000} is applied for feature extraction. 
Convolutional Gated Recurrent Unit (ConvGRU) \cite{shi2017deep} is employed to estimate the detection results based on the current frame and the previous states. by By estimating progress attributes, our action tubes are generated by a novel online connection method which computes the temporal boundary.
PR-RNN detector is evaluated on two unconstrained video datasets. Experimental results demonstrate that the progress attributes improve the action scoring and provide a better localization accuracy especially on the temporal domain. Our PR-RNN outperforms the state-of-the-art methods for most of Interaction over Union (IoU) thresholds in benchmark datasets. Our proposed PR-RNN provides an alternative way to model the temporal information without increasing the input length \cite{kalogeiton2017action,hou2017tube}, which also retains the online processing manner with the speed of 20 frame per second (fps).

In summary, our work makes the following contributions:
\begin{itemize}
    \item 	We introduce two new action attributes for spatial-temporal action localization: progression probability and progress rate.
    \item 	We build a novel RNN based on ConvGRU \cite{shi2017deep}, which takes in two-stream input, regresses the conventional outputs plus the two newly added attributes, and is able to meet the real-time requirement during online testing.
    \item 	We demonstrate that the proposed PR-RNN significantly improves the accuracy of localization and achieves the state-of-the-art performance for most of the IoU thresholds on two benchmark datasets.
\end{itemize}


\section{Related Work}\label{secRe}
CNN and Recurrent Neural Network (RNN) based action analysis methods have been extensively studied and achieved excellent results. Previous works are related to ours in three aspects: (1) temporal modeling for action representation; (2) object detection; and (3) spatial-temporal action localization.

\textbf{Action Representation.} Previously, to represent an action, handcrafted features \cite{dalal2005histograms,dalal2006human} are extracted densely \cite{wang2015action,wang2013action} or from spatial-temporal interest points \cite{bay2006surf} as local features and global features are obtained by encoding the local features with Bag-of-Words (BoW) \cite{sivic2003video} or Fisher vector \cite{perronnin2010improving}. Recently, researchers have developed quite a few effective frameworks for action analysis based on the CNN technique. 
There are three widely used strategies for action representation:

(i) 3D CNN based methods \cite{tran2015learning,carreira2017quo,ji20133d} inflate 2D convolutional filters with a temporal dimension, which is capable for generating representations from 3D receptive field straightforwardly. One issue with these architectures is that they have many more parameters and require much more computation resources to train the network due to the additional filter dimension. 

ii) Two-stream CNN based methods \cite{simonyan2014two,wang2016temporal,feichtenhofer2016convolutional} involve optical flow map as a new type of CNN input, which is helpful to capture low-level motion information of the actions. \cite{wang2016temporal,simonyan2014two} stack 10 optical flow maps from multiple consecutive frames. Differently, \cite{gkioxari2015finding} transform every single optical flow map into a 3-channel image, which is more efficient.

iii) RNN based methods \cite{yue2015beyond,donahue2015long,dave2017predictive} take convolutional features as the input of RNN layers, \textit{e.g.} Long Short Term Memory (LSTM) or Gated Recurrent Unit (GRU), to learn the temporal dependency among frame-level features. Shi \textit{et al.} propose convolutional Long Short Term Memory (ConvLSTM) and ConvGRU in \cite{shi2015convolutional,shi2017deep}, which replace the multiplication operation within RNN units by 2D convolution. It keeps the original spatial relationship in the feature maps while modeling the temporal dependency. 

To achieve efficient online action localization, in our proposed method, 2D CNN framework \cite{redmon2017yolo9000} with two-stream input is employed as the backbone network, followed by one ConvGRU \cite{shi2017deep} layer for temporal dependency modeling and localization results estimation. 


\textbf{CNN based object detection} Object detection is to localize the target objects in the images. Recently proposed methods for object detection \cite{girshick2014rich,girshick2015fast,ren2015faster,redmon2016you,liu2016ssd,redmon2017yolo9000}, which are built upon CNN, can be divided into two types: two-stage object detection and one-stage object detection. 
Two-stage object detection methods \cite{girshick2014rich,girshick2015fast,ren2015faster} first generate object proposals by distinguishing objects from the background on the predefined anchors, followed by object classification and bounding box regression for each proposal. 
Faster R-CNN \cite{ren2015faster} generates proposals by a Region Proposal Network (RPN). Features of the proposals are computed by a Region-of-Interest (RoI) pooling layer, which is used to regress the box and classify the object.
Differently, one-stage object detection methods \cite{redmon2016you,liu2016ssd,redmon2017yolo9000} simultaneously regress bounding boxes and classify the objects, which is more efficient. 
YOLO \cite{redmon2016you} divides the image into multiple cells and predicts two boxes in every cell. Features from the last convolution layer are used to regress the objectness score and bounding box and classify the object directly without RoI pooling. YOLOv2 \cite{redmon2017yolo9000} utilizes a Fully Convolutional Network (FCN) and introduces anchor box, where the network is trained to regress the offsets between anchor boxes and the ground truth.

\textbf{Spatial-temporal Action Localization.} Spatial-temporal action localization can be seen as the extension of object detection in the temporal domain, where the outputs are action tubes that consist of a sequence of bounding boxes. Some methods \cite{yu2015propagative,wang2014detecting,shao2014efficient} treat this task as a searching problem. Yu \textit{et al.} \cite{yu2015propagative} propose propagative Hough Voting to match the local features and propagate the label as well as the localization result. Some other methods \cite{weinzaepfel2015learning,soomro2015action,jain2014action,yu2015fast} model the task as a region proposal classification problem. 
Yu and Yuan \cite{yu2015fast} apply fast human and motion detectors \cite{dollar2014fast,benenson2012pedestrian} to compute bounding boxes, and then the tube-level proposals are generated by a maximum set coverage formulation.
Jain \textit{et al.} \cite{jain2014action} compute tubelets with hierarchical super-voxels as proposals, which are classified based on Dense Trajectory Feature (DTF) \cite{wang2011action}. 
Soomro \textit{et al.} \cite{soomro2015action} also use super-voxels to collect low-level cues, while action proposals are generated by 3D Conditional Random Field (CRF). 

Recently, the progress made by CNN based object detector inspires researchers to train action detectors with CNN \cite{saha2016deep,peng2016multi,kalogeiton2017action,hou2017tube,weinzaepfel2015learning,kalogeiton2017joint,saha2017amtnet,zolfaghari2017chained}. 
Gkioxari and Malik \cite{gkioxari2015finding} build a two-stream R-CNN framework to generate frame-level action proposal, which is further linked by dynamic programming. 
Peng and Schmid \cite{peng2016multi} extend Faster R-CNN \cite{ren2015faster} detector by a multi-region strategy, which extracts features from multiple regions of a proposal to improve the action proposal classification results. The action tubes are obtained by linking bounding boxes and temporal trimming. 
To improve the efficiency, Singh \textit{et al.} \cite{singh2016online} apply two-stream Single Shot Multibox (SSD) \cite{liu2016ssd} detector to estimate frame-level proposals and introduce an online Viterbi algorithm to link bounding boxes. Moreover, a fast optical flow map is employed during testing to achieve real-time testing speed. 
To further exploit temporal information of the action, some methods try to involve more information such as temporally expanding the input of CNN.
Zolfaghari \textit{et al.} \cite{zolfaghari2017chained} integrate the extracted human pose \cite{oliveira2016efficient} as a new stream of input, where multiple cues are added into the network successively by a Markov chain model. Kalogeiton 
\textit{et al.} \cite{kalogeiton2017action} temporally stack CNN features from multiple frames. The bounding boxes regression and action classification of multiple frames are processed simultaneously, which achieves better localization results than using a single frame. 
Different from multi-frame stacking, Hou \textit{et al.} \cite{hou2017tube} build a 3D CNN and propose the Tube-of-Interest (ToI) pooling method based on 3D convolution to generate action proposals. Li \textit{et al.} \cite{li2018recurrent} propose to improve the accuracy and stability of the action proposals by estimating the movement of the bounding boxes between two neighboring frames.

Our PR-RNN differs from the above mentioned methods as we focus more on output than input. In our work, two additional outputs, \textit{i.e.} progression and progress rate, are proposed to describe a bounding box, which learn the temporal dependency within actions in a supervised manner.

\section{Proposed Method}\label{secPR-RNN}
A progress regression method is proposed for exploiting more temporal attributes of an action. In this section, we first briefly present the original detector as our baseline model (Section \ref{secYOLO}). Then the proposed action progress regression method (Section \ref{secAPR}) and a novel action detector built on the progress regression mechanism (Section \ref{secDetector}) are introduced. Finally, the online action tube generation method (Section \ref{secTube}) is described.

\subsection{Baseline Action Detector}\label{secYOLO}
Action detectors input the extracted features and output several attributes of the action to build the final action localization result.
Previous works follow the framework of object detection, which only predict the label and the spatial position of an action. 
Taking one-stage detector YOLOv2 \cite{redmon2017yolo9000} as our baseline, it divides the input frame into $S\times S$ cells and estimates $B$ bounding boxes in each cell. Thus, the final prediction is a tensor with the size of $S\times S\times B\times(5+C)$, where $C$ is the number of action classes. One actionness score $s^{(A)}$, four coordinate offsets $(x,y,w,h)$, which are used to adjust the predefined anchor box, and $C$ classification probabilities $\{s^{(C)}_c\}_{c=1}^{C}$ are estimated to describe a bounding box.
The overall loss function of YOLOv2 detector can be expressed as:
\begin{equation}
\label{equ_ori_loss}
L_{YOLOv2}=\sum_{i=1}^{S^2}\sum_{j=1}^{B}L^{(coord)}_{ij}+L^{(conf)}_{ij}+L^{(cls)}_{ij}.
\end{equation}
$L^{(coord)}_{ij}$, $L^{(conf)}_{ij}$, and $L^{(cls)}_{ij}$ are the loss terms for coordinates, actionness score, and classification probabilities respectively: 
\begin{equation}
\label{equ_loss_coord}
\begin{split}
L^{(coord)}_{ij}=\lambda_{coord}\mathbbm{1}_{ij}^{act}&\Big[(x_{ij}-\hat{x}_{ij})^2+(y_{ij}-\hat{y}_{ij})^2\Big.\\+&\left.(w_{ij}-\hat{w}_{ij})^2+(h_{ij}-\hat{h}_{ij})^2\right],
\end{split}
\end{equation}
\begin{equation}
\label{equ_loss_conf}
L^{(conf)}_{ij}=\lambda_{act}\mathbbm{1}_{ij}^{act}(s^{(A)}_{ij}-1)^2+\lambda_{noact}\mathbbm{1}_{ij}^{noact}(s^{(A)}_{ij}-0)^2,
\end{equation}
\begin{equation}
\label{equ_loss_cls}
L^{(cls)}_{ij}=\lambda_{cls}\mathbbm{1}_{ij}^{act}\sum_{c=1}^{C}(s^{(C)}_{c,ij}-\hat{s}^{(C)}_{c,ij})^2,
\end{equation}
where $\mathbbm{1}_{ij}^{act}$ is an indicator function which equals to 1 if an action appears in cell $i$ and the $j$-th anchor box is responsible for this action. Similarly, $\mathbbm{1}_{ij}^{noact}$ equals to 1 if there are no actions. $\lambda_{coord}$, $\lambda_{act}$, and $\lambda_{cls}$ are weights of different components. 

\subsection{Action Progress Regression}\label{secAPR}
\begin{figure*}[ht]
    \begin{minipage}[b]{1\textwidth}
         \includegraphics[width=\textwidth]{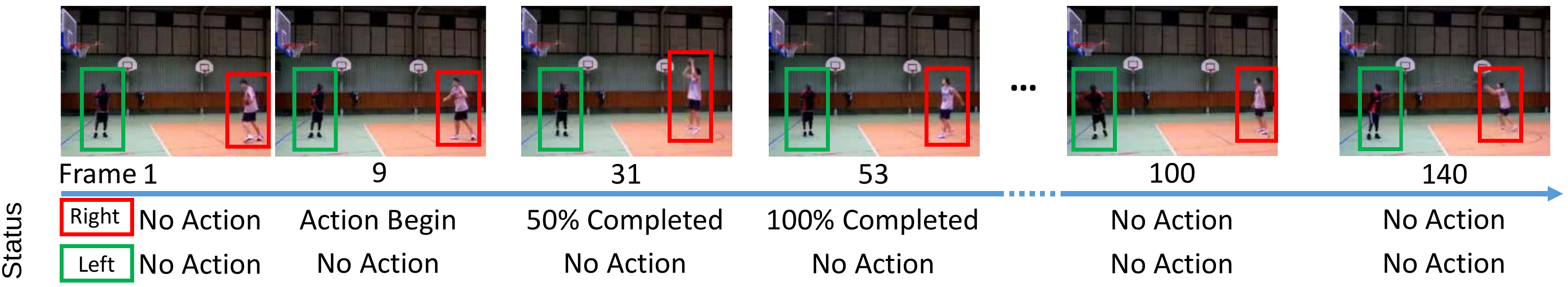}
        \centerline{(a) Action Temporal Status}\medskip \\
        \includegraphics[width=\textwidth]{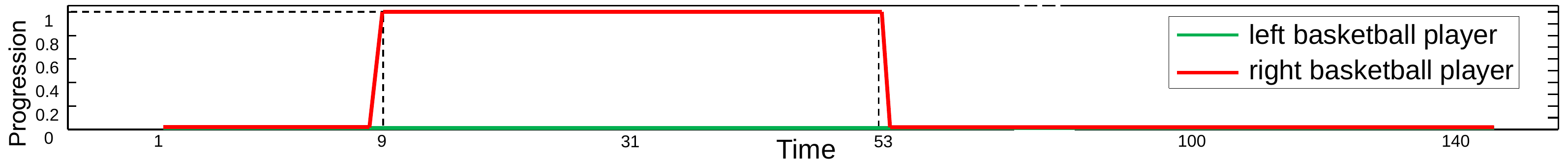}
        \centerline{(b) Ground truth of progression }\medskip\\
        \includegraphics[width=\textwidth]{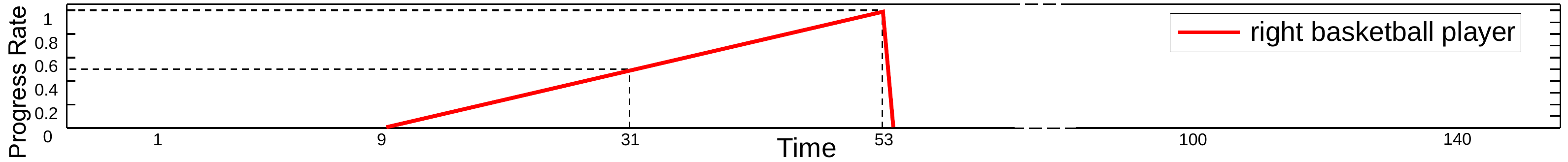}
        \centerline{(c) Ground truth of progress rate }\medskip 
    \end{minipage}
    \caption{Illustration of the action temporal status of ``basketball shooting'' of the two players. Top: temporal status shows whether the person is performing the action or not and the rate that the action has been performed. Middle: binary ground truth of progression indicates the temporal engagement; Bottom: continuous ground truth of progress rate represents the proportion of the action progress. }
    \label{FigStatus}
\end{figure*}
Previous action detectors are trained based only on the bounding box position and the action label. However, videos contain richer temporal information than static images. Given a frame of a video, we not only have the spatial position (bounding box) of the person, but also know the temporal status of the action at current time step. 
Temporal status includes two types of information: i) temporal engagement describes whether the person is performing a specific action; ii) temporal ratio tells the proportion that the action has been performed. Figure~\ref{FigStatus}(a) gives an example of the temporal status for ``basketball shooting''. If the action is not being performed, the status is ``no action''; otherwise it indicates the temporal rate. To quantize the temporal action status, our proposed action detector additionally estimates $C$ progression probabilities and $C$ progress rates, which represent the temporal engagement and the action progress rate of each action class respectively. 

\subsubsection{Progression} Progression describes the probability of a specific action being performed, which is denoted as $\{s^{(H)}_c\}_{c=1}^{C}$. As mentioned in the introduction, one actionness score is not enough to distinguish all the possible actions from backgrounds. Thus, in our method, the actionness score $s^{(A)}$ is tolerant to false positive results.
After the classification results $s^{(C)}_c$ divide the actionness score to $C$ action classes, the progression probability $s^{(H)}_c$ is introduced to predict the possibility of the $c$-th action in progress. Therefore, the final possibility for the $c$-th action in the bounding box $P(c|box)$ is computed by
\begin{equation}
\label{equ_final_score}
P(c|box)=s^{(A)}_{ij}\cdot s^{(C)}_{c,ij}\cdot s^{(H)}_{c,ij}.
\end{equation}
$s^{(H)}_{c}$ is the output of sigmoid activation function $\sigma(\cdot)$ independently instead of softmax function over all the action categories. Hence, the summation of progression probabilities of all the classes may not be 1. Progression regression in our model is seen as a re-scoring mechanism for each class, where some false positive detection results due to high actionness scores on irrelevant actions can be eliminated by suppressing the final confidence score. 

The ground truth of the progression for a cell is 1 if an action instance exists in that cell or 0 otherwise, as plotted in Figure \ref{FigStatus}(b). To train the progression regressor, boxes containing specific actions are selected as positive samples, \textit{i.e.}, $\mathbbm{1}_{c,ij}^{act}=1$, where $\mathbbm{1}_{cij}^{act}$ is an indicator function which equals to 1 if the $c$-th action appears in cell $i$ and the $j$-th anchor box is responsible for this action. Boxes are selected as negative samples if the box do not contain any actions, \textit{i.e.}, $\mathbbm{1}_{ij}^{noact}=1$ and the actionness score of the box is larger than a threshold $\theta$, \textit{i.e.}, $\mathbbm{1}_{ij}^{s^{(A)}>\theta}=1$. Therefore, the loss function for progression regression is defined as
\begin{equation}
\label{equ_loss_hp}
L^{(hp)}_{ij}=\sum_{c=1}^{C}\lambda_{hp}\mathbbm{1}_{c,ij}^{act}(s^{(H)}_{c,ij}-1)^2+\mathbbm{1}_{ij}^{noact}\mathbbm{1}_{ij}^{s^{(A)}>\theta}(s^{(H)}_{c,ij}-0)^2,
\end{equation}
where $\lambda_{hp}$ is the trade-off factor between positive and negative samples and the threshold $\theta$ is set to 0.2 in our network. 

\subsubsection{Progress Rate} Progress rate, denoted as $\{r_c\}_{c=1}^{C}$, is defined as the progress proportion that the action has been performed. Actions of one category may be performed with different speed, however, they follow a similar temporal procedure, such as ``run-jump-land'' for the action ``long jump''. Hence, the progress rate is a representative variable to describe the relative temporal position of a bounding box in an action tube. Progress rates provide an alternate way to model the temporal dependency of an action. If progress rates in a sequence of boxes are incremental, these boxes are more likely to contain an action. Moreover, the starting and ending locations of an action can also be inferred by the progress rate. That the score starts to increase from a low value indicates the beginning of an action, while that the score drops from a high value indicates the end of an action. An example is shown in Figure \ref{FigOverview}, where the progress rates of the golf player are incremental as he is performing the action of ``golf swing''. 

In recent years, some researchers also tried to explore the temporal dependency within an action \cite{yuan2017temporal,wang2016temporal}. All of these work manually divide an action into a pre-defined number of temporal states with fixed proportions of length of an action. This strategy usually trains a classifier to distinguish different temporal states of an action. However, the main drawback of these methods is that frames from different states, especially near the boundary of two neighboring states, are very similar and not easy to be classified correctly. Differently, we treat the temporal dependency modeling as a regression problem, which aims to estimate the exact relative temporal position of a frame in an action. In Section \ref{secExp}, the experimental results demonstrate that our model is capable of regressing gradually increasing progress rates of actions.

In an action tube with length $L$, the ground truth of progress rates for the $t$-th bounding box is set to $t/L$, $t=1\cdots L$, as shown in Figure \ref{FigStatus}(c). During training, the loss for progress rates is only computed for the box where the action appears. Thus the loss function is defined as
\begin{equation}
\label{equ_loss_sp}
L^{(sp)}_{ij}=\lambda_{sp}\mathbbm{1}_{ij}^{act}\sum_{c=1}^{C}(r_{c,ij}-\hat{r}_{c,ij})^2,
\end{equation}
where $\lambda_{sp}$ is the weight for the loss on progress rates.

\subsection{PR-RNN Detector}\label{secDetector}
By integrating the progression and progress rate regression into YOLOv2 action detector, we propose the PR-RNN action detector, which is capable of inferring rich temporal information of actions. The loss function of the proposed PR-RNN detector $L_{PR}(\cdot)$ is defined as the combination of the loss function of YOLOv2 and the two new progress regression components:
\begin{equation}
\label{equ_apr_loss}
L_{PR}=\sum_{i=1}^{S^2}\sum_{j=1}^{B}L^{(coord)}_{ij}+L^{(conf)}_{ij}+L^{(cls)}_{ij}+L^{(hp)}_{ij}+L^{(sp)}_{ij}.
\end{equation}
At each time step, the output of PR-RNN detector is a tensor with the size of $S\times S\times B\times(5+3C)$, where $S\times S\times B$ bounding boxes are regressed. For each box, $(5+3C)$ attributes are estimated, including an actionness score, 4 bounding box offsets, $C$ classification scores, $C$ progression scores, and $C$ progress rates. Then the final confidence scores are computed by Eq. \ref{equ_final_score}, which are denoted as $s_{c,ij}$. Non-maximum Suppression (NMS) is applied to eliminate the redundant boxes. Subsequently, action tubes are generated based on the predicted boxes and the corresponding attributes.

\subsection{Online Action Tube Generation}\label{secTube}
Without progress rates, action tubes are usually generated by linking the box with the highest score to the existing tube constrained by an IoU threshold \cite{singh2016online}, where the temporal relations among bounding boxes are not fully exploited. With the additional information of progress rates, we propose a novel action tube generation which takes temporal order into consideration and performs tube generation and temporal trimming in one online procedure. As a progress rate indicates the rate of an action has been performed, the proposed method aims to find a sequence of bounding boxes with high confidence scores and increasing progress rates in one online process. 

Action tubes are generated for every class separately. For the rest of this section, the tube generation method is discussed for one class, where the subscript is waived for simplicity. For a specific action class, the input of the tube generation method is a set of bounding boxes $B=\{\textbf{b}_{i}^{(t)}|\textbf{b}_{i}^{(t)}=(b_{i}^{(t)},s_{i}^{(t)},r_{i}^{(t)})\}_{i=1:n^{(t)},t=1:T}$. Each box contains a spatial position $b_{i}^{(t)}$, a confidence score $s_{i}^{(t)}$ and a progress rate $r_{i}^{(t)}$. The output is $M$ action tubes $\{(\{\hat{\textbf{b}}_m^{(t)}\},\{\hat{l}_m^{(t)}\},\bar{s}_m)\}_{t=t_m^{(s)}:t_m^{(e)},m=1:M}$. $\hat{l}_m^{(t)}$ is the corresponding temporal label sequence, which provides accurate temporal location of an action in the tube. $\bar{s}_m$, $t_m^{(s)}$, and $t_m^{(e)}$ are the average score of all boxes, starting and ending time step of the tube respectively. 
As estimated progress rates are noisy, to precisely detect the temporal location of an action in the testing video, we propose to use two variables $N_\uparrow$ and $N_\downarrow$ to accumulate the number of frames with increasing and decreasing progress rate for each tube. If the progress rate of the current frame is larger than the last frame, then the accumulation variables $N_\uparrow=N_\uparrow+1$ and $N_\downarrow=N_\downarrow-1$, and vice versa. The temporal part with increasing progress rates is detected if $N_\uparrow$ is larger than a threshold and vice versa, which is robust to the sudden change of progress rates.

\begin{algorithm}[t]                    
    \caption{Online Temporal Labeling in the Action Tube }          
    \label{alg_1}                           
    \begin{algorithmic}[1]                    
        \REQUIRE $\{\hat{\textbf{b}}_{m}^{(\tau)}\}_{\tau=t_m^{(s)}:t}$, $\{\hat{l}_{m}^{(\tau)}\}_{\tau=t_m^{(s)}:t}$, $N_\uparrow$, $N_\downarrow$, $\alpha$, $K$;
        
        \ENSURE $\{\hat{l}_{m}^{(\tau)}\}_{\tau=t_m^{(s)}:t}$, $N_\uparrow$, $N_\downarrow$;
        
        \STATE INITIALIZE $\hat{l}_m^{(t)}=\hat{l}_m^{(t-1)}$;
        
        \IF{$\hat{r}_m^{(t)}>\hat{r}_m^{(t-1)}$} 
        \STATE $N_\uparrow=\min(K,N_\uparrow+1)$, $N_\downarrow=\max(0,N_\downarrow-1)$
        \ELSE
        \STATE $N_\downarrow=\min(K,N_\downarrow+1)$, $N_\uparrow=\max(0,N_\uparrow-1)$
        \ENDIF 
        
        \IF{$N_\uparrow=K$}
        \STATE UPDATE $\{\hat{l}_m^{(\tau)}\}_{\tau=t-K+1:t}=1$
        \ELSIF{$N_\downarrow=K$}
        \STATE UPDATE $\{\hat{l}_m^{(\tau)}\}_{\tau=t-K+1:t}=0$
        \ELSIF{$\hat{s}_{m}^{(t-K+1:t)}>\alpha$}
        \STATE  ADJUST $\hat{l}_m^{(t-K+1:t)}=1$
        \ENDIF        
    \end{algorithmic}
\end{algorithm}

\begin{figure*}[t]
    \centering
    \includegraphics[width=1\textwidth]{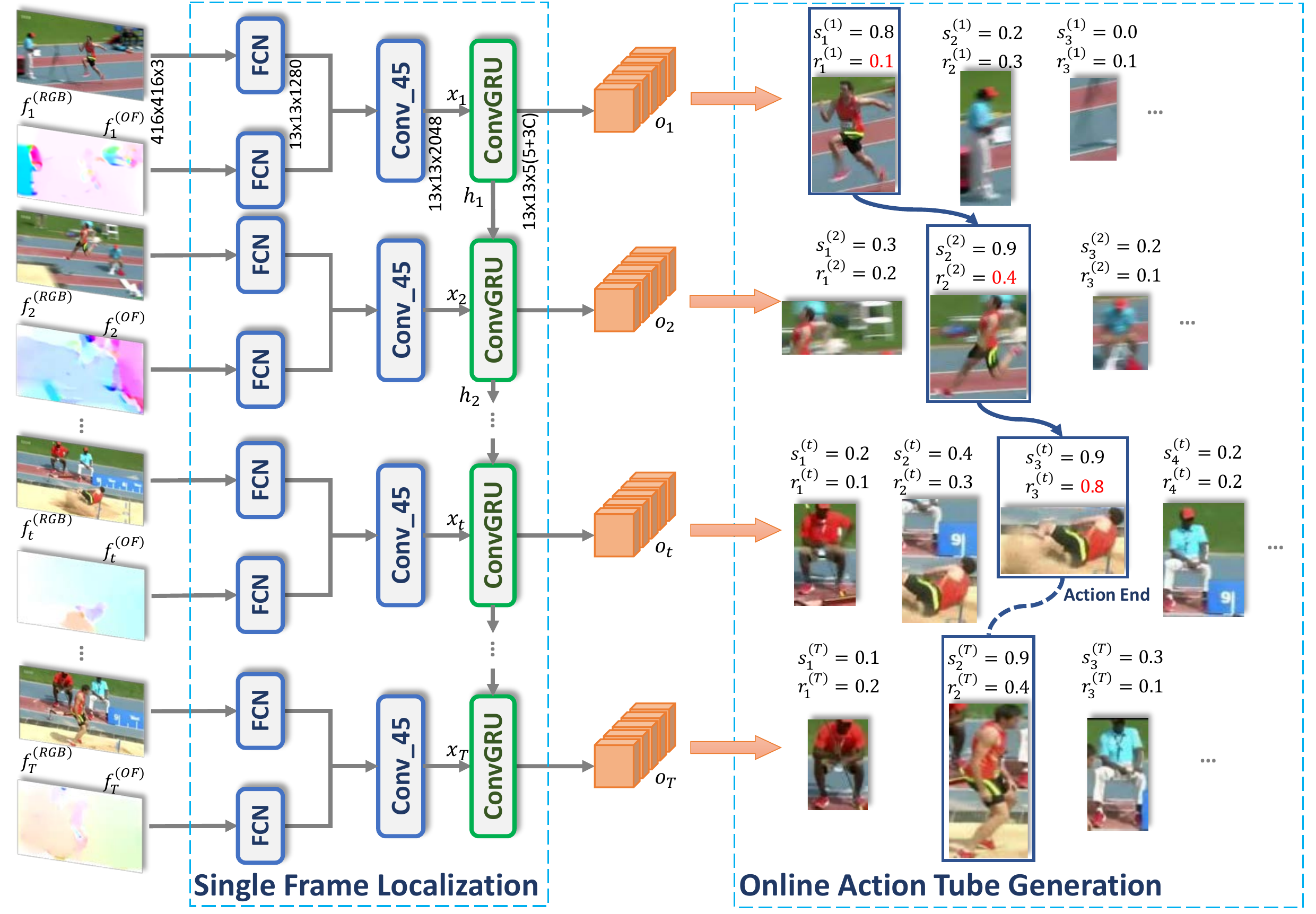}
    \caption{The overview of the proposed PR-RNN. 1) The architecture of PR-RNN is illustrated in the left of the figure. RGB frame and optical flow map are processed separately and stacked before the last convolution layer (Conv\_45). ConvGRU infers the detection results based on convolutional feature maps from the current frame and hidden state feature maps from the previous frame. 2) The novel online action tube generation method is shown in the right of the figure. During the online tube generation step, the model aims to find a bounding box sequence with high confidence scores as well as increasing progress rates }
    \vspace{-1em}
    \label{FigArch}
\end{figure*}

To link an action tube and estimate the temporal label simultaneously, the following steps are applied:
\begin{enumerate}
    \item $t=1$, initialize $M$ tubes by finding $M$ best boxes with the highest score from $B^{(1)}=\{\textbf{b}_{i}^{(1)}\}_{i=1:n^{(1)}}$. The initial label $\hat{l}_m^{(1)}=0$. $N_\uparrow$ and $N_\downarrow$ are initialized to 0.
    \item Traverse all video frames from $t=2$ to $t=T$, execute steps (a) to (c) in each frame.
    \begin{enumerate}
        \item Sort existing tubes by $\bar{s}_m$ in descend order and keep the first $M$ tubes.
        \item Traverse all tubes from $m=1$ to $m=M$. Execute steps (i) to (v) for each tube.
        \begin{enumerate}
            \item If the tube is not completed, build a subset of boxes $B_m^{(t)}=\{\textbf{b}_{i}^{(t)}\in B^{(t)}|IoU(b_{i}^{(t)},\hat{b}_{m}^{(t-1)})>\gamma\}_{i=1:n^{(t)}}$.
            \item If $B_m^{(t)}\neq \emptyset$, link the box $\textbf{b}_{j}^{(t)}$, which has the highest score in $B_m^{(t)}$, to the $m$-th tube. 
            \item Update the bounding box set $B^{(t)}=B^{(t)}\backslash\textbf{b}_{j}^{(t)}$.
            \item Update the average score of the $m$-th tube $\bar{s}_m=avg(\{\hat{s}_{m}^{(\tau)}\}_{\tau=t_m^{(s)}:t})$.
            \item Compute temporal labels $\hat{l}_m^{(t)}$ based on $\{\hat{r}_{m}^{(\tau)}\}_{\tau=t_m^{(s)}:t}$, $\{\hat{s}_{m}^{(\tau)}\}_{\tau=t_m^{(s)}:t}$ and accumulators $N_\uparrow$ and $N_\downarrow$. The procedure of temporal labeling in an tube is summarized in Algorithm \ref{alg_1}.
            \item Complete the $m$-th tube, if the tube is not linked in the recent $K$ frames.
        \end{enumerate}
        \item Traverse all the rest boxes in $B^{(t)}$ from $i=1$ to $i=\|B^{(t)}\|$, start a new tube.
    \end{enumerate}
\end{enumerate}

The threshold $\alpha$ in step 11 in Algorithm \ref{alg_1} is a trade-off factor to balance the effect of the confidence score and the progress rate. If $\alpha=1$, then only the progress rate affects the temporal labeling, where tubes are generated only by dependency between boxes. On the contrary, the progress rate mechanism will be disabled if $\alpha=0$, where the tubes are linked only based on the confidence score of every single box. The selection of $\alpha$ will be discussed in Section \ref{secTest}. The final action tubes are obtained by further trimming the $M$ tubes according to the corresponding temporal labels. An example is given in the right of Figure \ref{FigArch}. In the last frame, the second bounding box is not linked into the action tube even its confidence score is high, which is represented by the dash line. This is because its estimated progress rate is decreasing, which indicates that the action has ended.

\section{Implementation Details}\label{secImplement}
In this section, the details of implementing our proposed PR-RNN are introduced, which contain the description of our network architecture and the detailed information in training and testing stages.

\subsection{Architecture of PR-RNN}\label{secNetwork}
In the proposed PR-RNN, two FCNs with the same architecture are utilized to extract features from RGB frame and optical flow map separately. The input size of two streams are both $416\times416\times3$, where we transform every single optical flow map into a 3-channel image. The architecture of FCNs follows YOLOv2, which is shown in the left of Figure \ref{FigArch}. The FCN has 22 convolutional layers, 5 max pooling layers and a passthrough layer that combines feature maps with different resolution. 
We fuse two FCNs by concatenating the feature maps before the Conv\_45. 
Furthermore, the number of filters in Conv\_45 is set to 2048 instead of 1024 in YOLOv2 as we generate more outputs. 
The output feature maps of Conv\_45 is sent to the ConvGRU \cite{shi2017deep}, which applies $B\times(5+3C)$ convolutional filters with the size of $3\times3$ at every gate. To suit our detection task, different activation functions are employed in our output gate of the ConvGRU layer. We replace the $\tanh(\cdot)$ activation functions by $\sigma(\cdot)$ for actionness, bounding box coordinate offsets, progression and progress rate regression, replace the $\tanh(\cdot)$ activation functions by $softmax(\cdot)$ for action classification, and remove the activation functions of bounding box width and height offsets regression.

\subsection{Network Training}\label{secTrain}
We use ImageNet pre-trained weights on FCNs for both of the two streams. Data augmentation, such as random rescaling, cropping, and flipping, is applied, which follows the training procedure of YOLOv2. 
The number of anchor boxes is set to $B=5$, where widths and heights are obtained by dimension clustering in \cite{redmon2017yolo9000}. To balance the training loss from all components, we set the trade-off factor of actionness $\lambda_{act}=10$, while other factors are set to 5. Due to the limitation of computing resource, FCNs and the ConvGRU layer are optimized separately. 
First, an additional convolutional layer with $5\times(5+3C)$ $1\times1$ convolutional filters is concatenated after Conv\_45 to train the two FCNs and Conv\_45 for 40 epochs by Eq. \ref{equ_apr_loss}, where the batch size is set to 20. 
The initial learning rate is $10^{-4}$, which decays 0.5 after 5, 10, and 20 epochs. 
Afterwards, we fix the weights of two FCNs and train the Conv\_45 ConvGRU layer in PR-RNN for 40 epochs, which takes 10-frame clips with a batch size of 20 as input.
The learning rate and decay scheme remain the same. It's worth mentioning that the estimation of the proposed progress rates heavily relies on the information from the previous frames. Thus, the first hidden state of a 10-frame clip should be initialized by the last hidden state of the previous clip.

\subsection{Online Detection}\label{secTest}
During testing, all input images are padded to $416\times416$ by 0 without rescaling. The whole video is sent into ConvGRU without cutting to clips. Estimated boxes whose confidence score larger than $10^{-3}$ are selected for tube generation. Note that there are many periodic actions defined as the action is repeated multiple times in an action tube, such as ``cycling'' and ``fencing''. For these periodic actions, the progress rates can hardly be predicted accurately since there are arbitrary periods in an action tube and all frames in a period are similar.
Ideally, for such a case, the confidence score should dominate the decision for temporal labeling. Thus, we propose to use the average training error $\epsilon$ to distinguish periodic actions and non-periodic actions, which is obtained by averaging the progress rates training errors of all positive samples within one class. Then a class specific trade-off factor $\alpha$ in Algorithm \ref{alg_1} is defined as $\alpha_c=\exp{(-\epsilon_c^2/10^{-2}})$. The confidence score of a bounding box being higher than the threshold, would mean sufficient confidence in the classification, where the box is then linked into a tube instead of further considering the progress rates. This strategy is simple but effective in detecting both periodic and non-periodic actions. We set $\gamma=0.3$, $K=6$ for online tube generation in all the experiments.

\section{Experiments}\label{secExp}

Extensive experiments are designed in this section to verify the effectiveness of the proposed PR-RNN. First, we present the information of the two datasets and the evaluation metrics we use. Then our proposed PR-RNN is evaluated on these datasets and the results are compared to the state-of-the-art methods.

\subsection{Datasets} 
As the proposed PR-RNN aims to improve the accuracy of temporal position of action tubes in spatial-temporal action localization, unconstrained videos, which contain temporal background, are required to evaluate our method. Hence, two action localization datasets are tested: UCF-101 \cite{soomro2012ucf101} and THUMOS'14 \cite{jiang2014thumos}.

\textbf{UCF-101} contains 24 action classes and more than 3000 videos for spatial-temporal action localization. The spatial-temporal positions of actions are annotated for a 24-class subset in \cite{jiang2014thumos}. 
There are 3 different training and testing splits provided in the dataset. Following the same setting as other methods, only the first split is tested in our experiment.

\textbf{THUMOS'14} consists of 1010 long unconstrained videos for action recognition and temporal localization. \cite{sultani2016if} provides spatial annotations for ``golf swing'' and ``tennis swing''. We use the model trained on split 2 and 3 of UCF-101 directly to test these two classes.

\subsection{Evaluation Metrics}
We evaluate the performance by frame-level mean Average Precision (f-mAP) and video-level mAP (v-mAP). the f-mAP is measured with a fixed IoU threshold 0.5, denoted as f-0.5. When measuring v-mAP, the overlap between two action tubes, denoted as tube-IoU, is obtained by multiplying the average spatial IoU in each frame and the temporal IoU (t-IoU). Multiple tube-IoU thresholds are evaluated, including an average performance between threshold 0.5 and 0.95 with a step of 0.05, which is denoted as 0.5:0.95.

\subsection{Ablation Study}

\textbf{Baseline (YOLOv2+ConvGRU).} Our baseline method is two-stream input YOLOv2 \cite{redmon2017yolo9000} with ConvGRU \cite{shi2017deep}, which is trained by Eq. \ref{equ_ori_loss} and predicts the same outputs as YOLOv2. Online linking without temporal labeling is applied, where the action tubes are only generated by confidence scores.

\begin{figure}[t]
    \centering
    \includegraphics[width=0.48\textwidth]{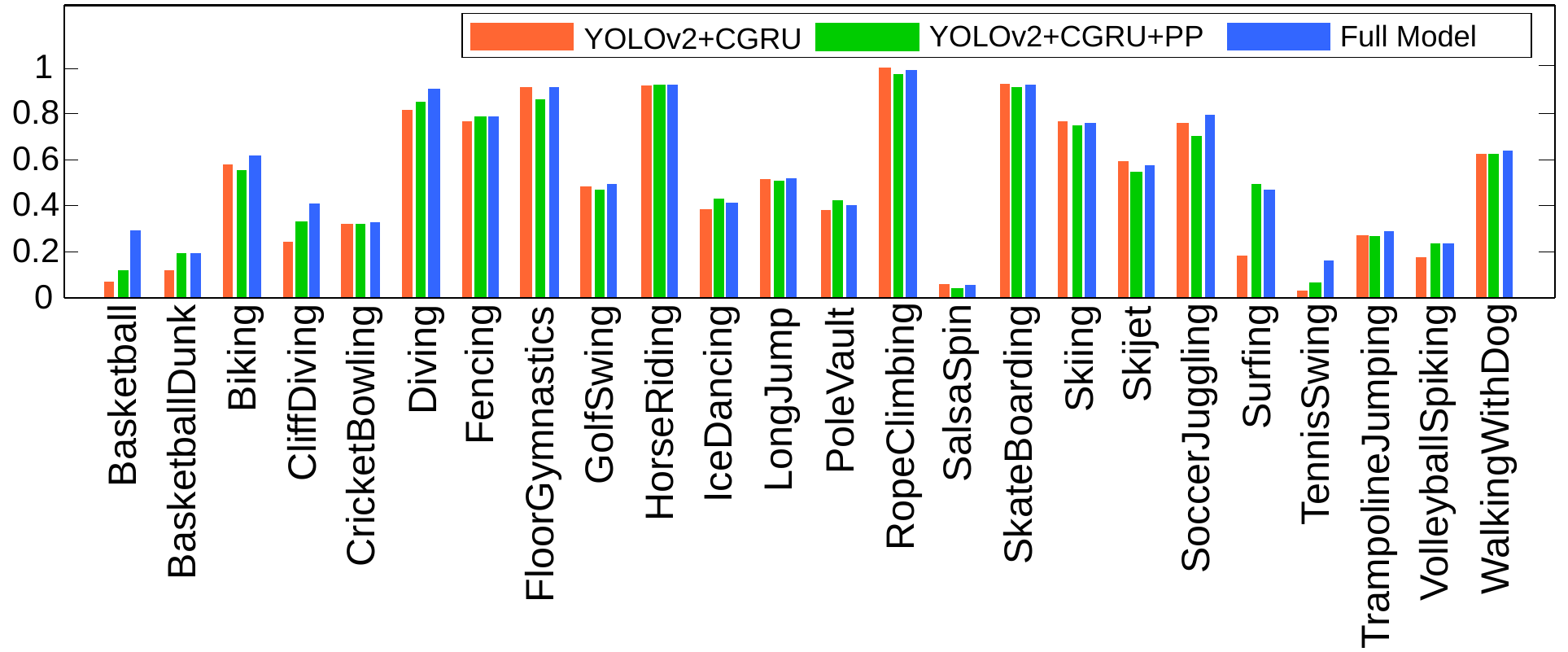}
    \caption{Video-level AP at $\delta=0.5$ in UCF-101. }
    \label{FigMap}
\end{figure}

\begin{figure}[t]
    \centering
    \includegraphics[width=0.48\textwidth]{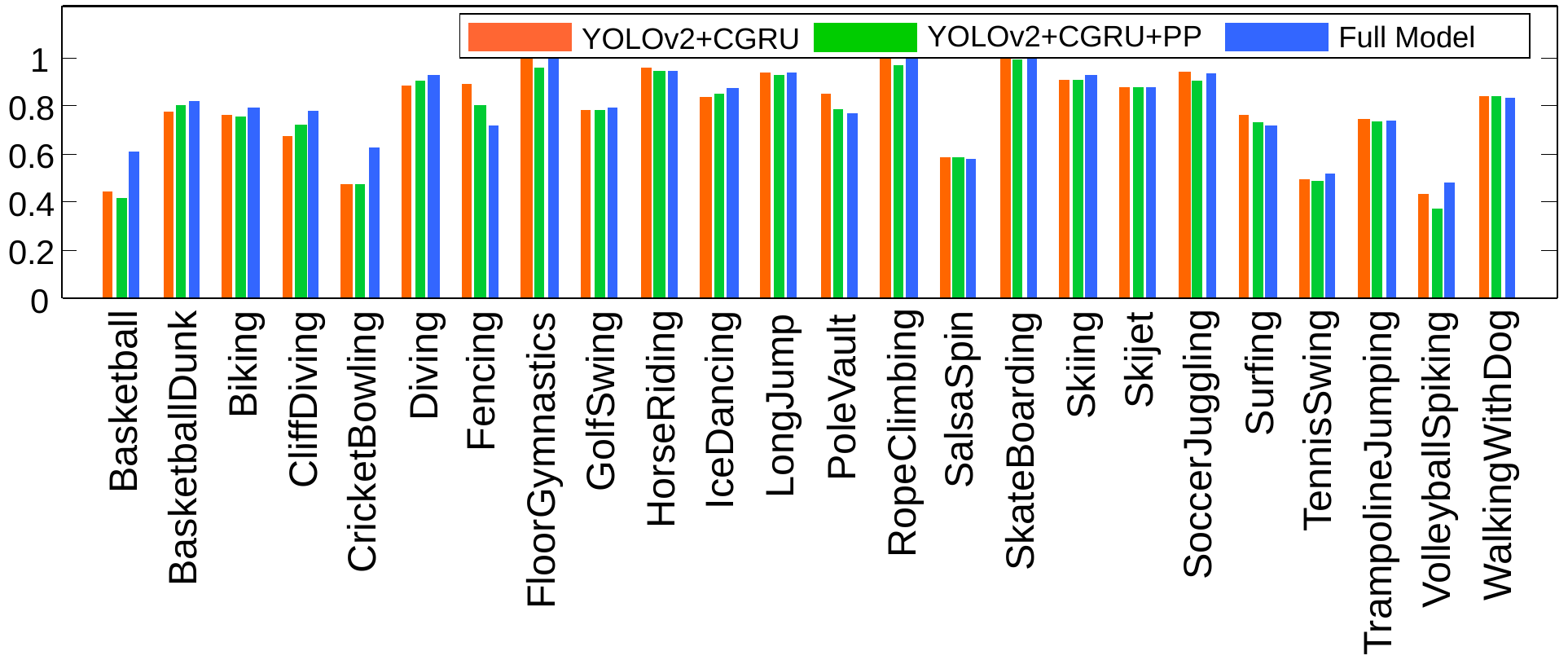}
    \caption{Average t-IoU of all action classes in UCF-101. }
    \label{FigIou}
\end{figure}
\begin{table}[ht]
    \renewcommand\arraystretch{1.1}
    \label{tab_ucf101}
    \caption{Comparisons to the baselines and state-of-the-art methods on UCF-101. The results of v-mAP with different tube-IoU thresholds are reported.} 
    \begin{center}
        \scriptsize
        \begin{tabular*}{0.48\textwidth}{@{\extracolsep{\fill}}l c  c  c  c  c  c  }
            \toprule[0.75pt]  
            IoU Threshold $\delta$ & 0.1 & 0.2 & 0.3 & 0.5 & 0.75 & 0.5:0.95 \\
            \toprule[0.75pt]  
            Weinzaepfel et al. 2015   & 51.7 & 46.8 & 39.2 & - & - & -  \\
            Peng and Schmid 2016      & 50.4 & 42.3 & 32.7 & - & - & - \\
            Zolfaghari et al. 2017    & 59.5 & 47.6 & 38.0 & - & - & -   \\
            Hou et al. 2017           & 51.3 & 47.1 & 39.2 & - & - & - \\
            Saha et al. 2016          & 76.6 & 66.8 & 55.5 & 35.9 & 07.9 & 14.4 \\
            Singh et al. 2017         &   -  & 73.5 &   -  & 46.3 & \textbf{15.0} & 20.4 \\
            Li et al. 2018            & 81.3 & \textbf{77.9} & \textbf{71.4} & - & - & -\\
            Kalogeiton et al. 2017    &   -  & 77.2 &   -  & \textbf{51.4} & \textbf{22.7} & \textbf{25.0} \\
            \hline
            YOLOv2+ConvGRU            & 79.4 & 74.6 & 66.6 & 48.4 & 11.0 & 18.9 \\
            YOLOv2+ConvGRU+PP         & \textbf{81.4} & 77.1 & 69.4 & 49.8 & 12.6 & 19.8 \\
             Full Model                & \textbf{82.3} & \textbf{78.0} & \textbf{69.8} & \textbf{54.1} & \textbf{15.0} & \textbf{22.8}  \\
            \bottomrule[0.75pt]
        \end{tabular*} 
    \end{center}
\end{table}

\begin{table}[ht]
    \renewcommand\arraystretch{1.1}
    \label{tab_frame}
    \caption{Comparisons to baseline methods on THUMOS'14. The results of v-mAP with different tube-IoU thresholds are reported.}
    \setlength{\belowcaptionskip}{ 10pt }
    \begin{center}
        \scriptsize
        \begin{tabular*}{0.35\textwidth}{@{\extracolsep{\fill}}l   c   c   c }
            \toprule[0.75pt]  
            IoU Threshold $\delta$ & 0.2  & 0.3  & 0.5  \\
            \toprule[0.75pt]  
            YOLOv2+ConvGRU         & 28.4 & 8.6  & 0.8  \\
            YOLOv2+ConvGRU+PP      & 30.3 & 14.6 & 1.0  \\
            Full Model             & \textbf{31.9} & \textbf{18.4} & \textbf{3.9}  \\
            \bottomrule[0.75pt]
        \end{tabular*} 
    \end{center}
\end{table}

\begin{table}[ht]
    \renewcommand\arraystretch{1.1}
    \label{tab_fmap}
    \caption{Comparisons to the baselines and state-of-the-art methods on both UCF-101 and THUMOS'14. The reuslts of f-mAP on spatial IoU threshold $\delta=0.5$ are reported}.
    \begin{center}
        \scriptsize
        \begin{tabular*}{0.35\textwidth}{@{\extracolsep{\fill}}l c c }
            \toprule[0.75pt]  
            f-0.5 & UCF-101 & THUMOS'14 \\
            \toprule[0.75pt]  
            Weinzaepfel et al. 2015   & 35.8 & - \\
            Peng and Schmid 2016      & 39.6 & - \\
            Hou et al. 2017           & 41.4 & - \\
            Kalogeiton et al. 2017    & \textbf{67.1} & - \\
            \hline
            YOLOv2+ConvGRU            & 65.4 & 26.9 \\
            YOLOv2+ConvGRU+PP         & \textbf{66.8} & \textbf{35.8} \\
            Full Model                & \textbf{66.8} & \textbf{35.8} \\
            \bottomrule[0.75pt]
        \end{tabular*} 
    \end{center}
\end{table}

\textbf{YOLOv2+ConvGRU+PP.} The effectiveness of the progression probability regression is first evaluated, where the method is denoted as ``YOLOv2+ConvGRU+PP''. In this method, bounding boxes are re-scored by Eq. \ref{equ_final_score} and action tubes are generated without progress rates as the baseline method. The f-mAP of spatial IoU threshold $\delta=0.5$ on UCF-101 is shown in the first column in Table III. By employing progression probabilities, the f-mAP is improved by around 1.4\% over the baseline. As progression does not change the number of detections but the confidence score of each detection, the gain on f-mAP is caused by suppressing the score of false positive detections. From Table I it can be observed that the proposed progression probability also achieve improvements in v-mAP, such as a gain of 1.4\% at the tube-IoU threshold of $\delta=0.5$. In THUMOS'14, the proposed progression improves the f-mAP by 8.9\% (see Table III), as the confidence scores of irrelevant actions, such as movings between two ``tennis swing'', are suppressed effectively. The performance gain in THUMOS'14 is larger than that in UCF-101 since THUMOS'14 has more long unconstrained videos than UCF-101, where suppressing the score of irrelevant actions is more effective.

\textbf{Full Model (YOLOv2+ConvGRU+PP+PR).} Our full model integrates both the progression and progress rate regression and applies the online tube generation with temporal labeling. Table I shows the results of v-mAP on UCF-101, the full model outperforms YOLOv2+ConvGRU+PP by 4.3\% in v-mAP with $\delta=0.5$ due to the effective temporal labeling with progress rates. 
Figure \ref{FigMap} depicts the video-level Average Precisions (v-AP) at $\delta=0.5$ of all classes in UCF-101, where our method achieves higher AP on most classes, especially on the non-periodic action classes, such as ``basketball'' and ``tennis swing''. For some non-periodic actions, such as ``long jump'', our performance gain is not significant because testing videos of these classes are trimmed already. 
The results on THUMOS'14 listed in Table II further demonstrate that our full model can achieve superior v-mAP on long unconstrained videos. For instance, the v-AP at $\delta=0.5$ of our detector surpasses the baseline by 5.5\% for the action ``golf swing''. 
The f-mAP of the full model on UCF-101 and THUMOS'14 is shown in Table III, where the f-mAP at $\delta=0.5$ is the same as that of YOLOv2+ConvGRU+PP, as progress rate has no effects on single bounding box scoring.

\begin{figure*}[t]
    \centering
    \includegraphics[width=1\textwidth]{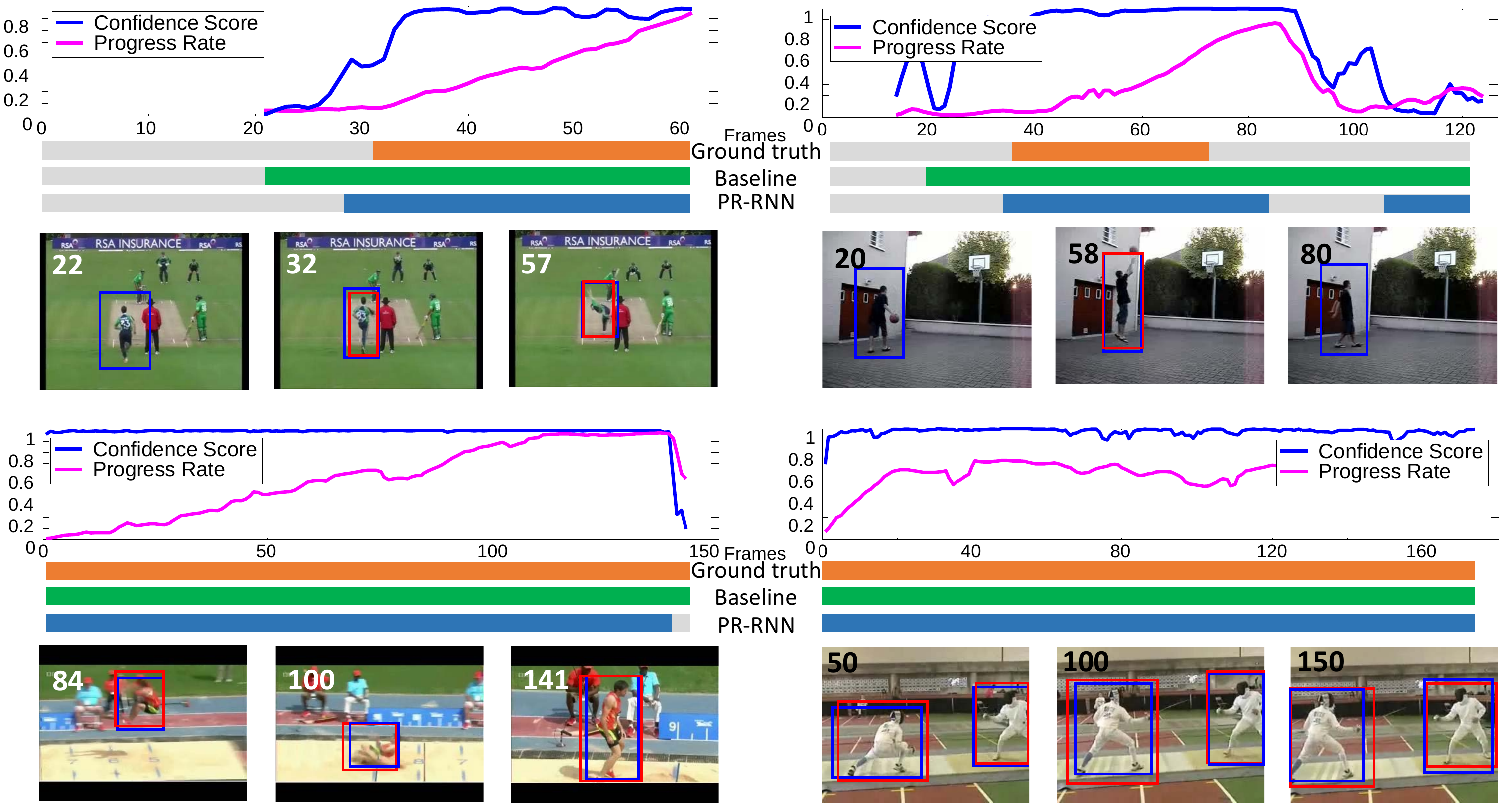}
    \caption{Examples of our detection results on four videos from UCF-101. Confidence scores and progress rates are represented by blue and pink curves. Color bars show the temporal localization results. Spatial localization results of three frames for each video are visualized, where the frame index is at the top-left conner of each frame. Red boxes are ground truths and blue boxes are from our PR-RNN. }
    \label{FigRes}
\end{figure*}

To further evaluate the temporal localization capability of the proposed method, average t-IoU is computed by averaging the t-IoU between the ground truth and the best estimated tube on UCF-101. The results of average t-IoU is shown in Figure \ref{FigIou}, which shows that our proposed progress rate and online temporal labeling improves the temporal localization accuracy for most of the non-period actions, such as ``basketball shooting'' and ``cricket bowling''. For periodic actions or actions in constrained videos, our detector and the baseline methods provide similar results. Some examples are visualized in Figure \ref{FigRes}, where the confidence score sequences is shown by the curves and the temporal localization results of different methods is represented by bars of different colors. In Figure \ref{FigRes} we can observe that the progress rate is predicted accurately for the first three non-periodic actions ``cricket bowling'', ``basketball shooting'', and ``long jump''. A periodic action ``fencing'' is also shown in the bottom right of Figure \ref{FigRes}. With the help of ConvGRU, our model also provides a sequence of increasing progress rates at the beginning, however, the estimated progress rates become unreasonable quickly, as the duration of the action is arbitrary and unpredictable. For these actions, confidence scores contribute more than progress rates on tube generation. 

With a single GPU (Nvidia Titan Xp), our online processing speed is 20 fps (optical flow computing is excluded) when the input size is $416\times416$ for both RGB image and optical flow map. For reference, the speed of original YOLOv2 network with single stream input is 33 fps with the same setting.

\subsection{Comparisons to the State-of-the-art}
We compare our PR-RNN detector to several state-of-the-art methods \cite{peng2016multi,zolfaghari2017chained,hou2017tube,saha2016deep,singh2016online,kalogeiton2017action,li2018recurrent} 
only on UCF-101, as these methods did not report any spatial-temporal action localization results on THUMOS'14. For \cite{peng2016multi}, the results with the multi-region scheme are reported. From the results in Table I and Table III, we can see that our detector achieves the state-of-the-art or the second best performance compared with all other methods. Our proposed method significantly outperforms the human pose based method \cite{zolfaghari2017chained} and R-CNN based methods \cite{peng2016multi,saha2016deep,hou2017tube} at all IoU thresholds. For instance, our method surpasses \cite{saha2016deep} by 18.2\% at $\delta=0.5$ and \cite{hou2017tube} by 30.3\% at $\delta=0.3$. The f-mAP of our method is slightly lower than the proposal based method \cite{li2018recurrent}, as it employs ResNet-101 \cite{he2016deep} as their backbone network, which is much powerful than YOLOv2. Moreover, \cite{li2018recurrent} links and trims the action tubes in an offline process.
Compared to the SSD based methods \cite{singh2016online,kalogeiton2017action}, our detector achieves supreme performance on f-mAP and v-mAP at the thresholds ranging from $\delta=0.1$ to $\delta=0.5$. Our detector outperforms the ACT-detector \cite{kalogeiton2017action} by 2.7\% and online SSD \cite{singh2016online} by 7.8\% when $\delta=0.5$. 
At the threshold $\delta=0.75$ and the average threshold $\delta=0.5:0.95$, our method also provides the second best v-mAP and achieves a comparable performance to ACT-detector. 
This is because the actions in every single-frame is estimated multiple times by ACT-detector and the final results are obtained by averaging multiple estimations from frame stacks, where the estimated bounding boxes are more accurate in the spatial domain. Furthermore, its temporal smoothing strategy is an offline procedure. Different from ACT-detector, our detector follows the online setting, \textit{i.e.}, (1) the detector makes the decision without future frames; (2) the history detection results should not be changed. The spatial accuracy affects our performance at the highest tube-IoU threshold, as tube-IoU is computed by multiplying the spatial IoU and the temporal IoU.
In summary, compared to these methods, our PR-RNN action detector benefits from the RNN based progression and progress rate regression, which infers the temporal status of an action and estimates more precise action tubes in the temporal domain, and achieves the stat-of-the-art performance on most of the IoU thresholds.

\section{Conclusions}\label{secCon}
We have proposed the Progress Regression RNN (PR-RNN) detector for online spatial-temporal action localization in unconstrained videos. Compared with the previous action detections, our proposed action detector predicts two extra attributes of actions: the progression probability and the progress rate. The progression probability can help eliminate false positive localization results by re-scoring confidence scores of bounding boxes. The progress rate learns the temporal dependency of an action in a supervised manner, which is further integrated with the online tube generation. The extensive experiments demonstrate that by introducing the progression probability and the progress rate, our detector estimates temporally more accurate action tubes. Our detector achieves the state-of-the-art performance for most of the IoU thresholds on the two benchmark datasets.	

\ifCLASSOPTIONcaptionsoff
  \newpage
\fi



%
%
%
\bibliographystyle{IEEEtran}
\bibliography{egbib}

%
\begin{IEEEbiography}[{\includegraphics[width=1in,height=1.25in,clip,keepaspectratio]{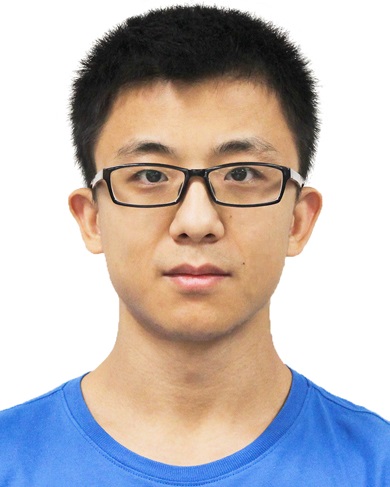}}]{Bo Hu} received the M.Eng. degree from Nanyang Technological University in 2017. He is now a Research Associate with the Institute for Media Innovation, Nanyang Technological University, Singapore. His research interests mainly include human action analysis in videos, computer vision, and machine learning.

\end{IEEEbiography}


\begin{IEEEbiography}[{\includegraphics[width=1in,height=1.25in,clip,keepaspectratio]{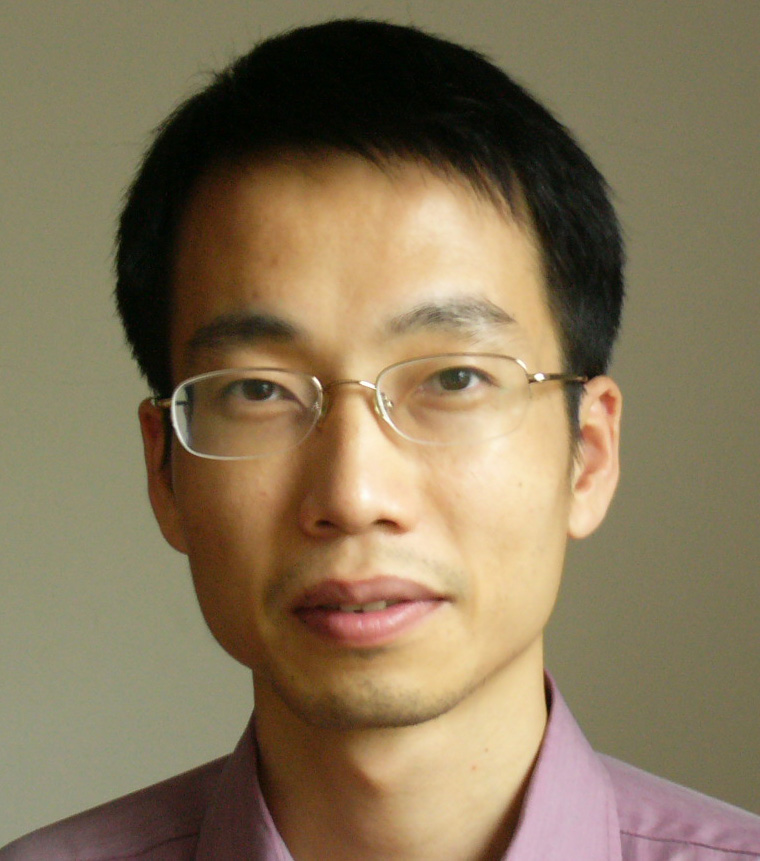}}]{Jianfei Cai}(S’98-M’02-SM’07) received his PhD degree from the University of Missouri-Columbia.
He is a full Professor and currently a Cluster Deputy Director of Data Science \& AI Research Center (DSAIR) at Nanyang Technological University (NTU), Singapore. He has served as the Head of Visual \& Interactive Computing Division and the Head of Computer Communication Division at NTU. His major research interests include computer vision, multimedia and deep learning. He has published over 200 technical papers in international journals and conferences. He is currently an Associate Editor for IEEE Trans. on Multimedia, and has served as an Associate Editor for IEEE Trans. on Image Processing and Trans. on Circuit and Systems for Video Technology.

\end{IEEEbiography}

\begin{IEEEbiography}[{\includegraphics[width=1in,height=1.25in,clip,keepaspectratio]{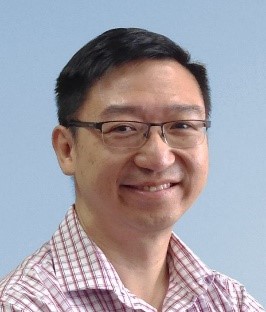}}]{Tat-Jen Cham} is an Associate Professor in the School of Computer Science \& Engineering, Nanyang Technological University, Singapore. After receiving his BA and PhD from the University of Cambridge, he was subsequently a Jesus College Research Fellow, and later a research scientist in the DEC/Compaq Research Lab in Cambridge, MA.
    Tat-Jen received overall best paper prizes at PROCAMS2005, ECCV1996 and BMVC1994, and is an inventor on eight patents. He has served as an editorial board member for IJCV, a General Chair for ACCV2014, and Area Chair for past ICCVs and ACCVs. 
    Tat-Jen’s research interests are broadly in computer vision and machine learning, and he is currently a co-PI in the NRF BeingTogether Centre (BTC) on 3D Telepresence.

\end{IEEEbiography}

\begin{IEEEbiography}[{\includegraphics[width=1in,height=1.25in,clip]{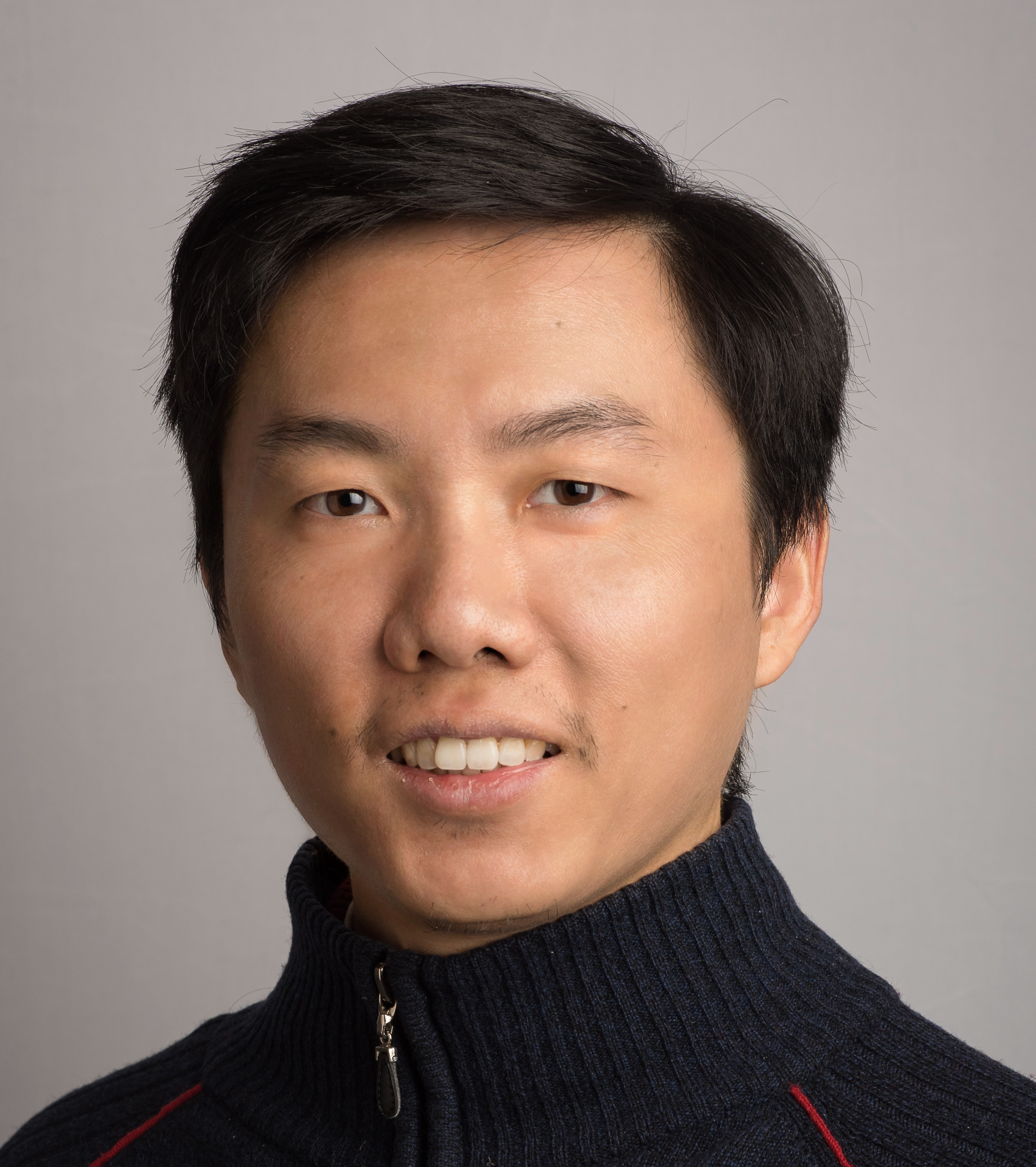}}]{Junsong Yuan} is currently an Associate Professor and Director of Visual Computing Lab at Department of Computer Science and Engineering (CSE), State University of New York at Buffalo, USA. Before that he was an Associate Professor at Nanyang Technological University (NTU), Singapore. He obtained his Ph.D. from Northwestern University, M.Eng. from National University of Singapore and B.Eng from the Special Program for the Gifted Young of Huazhong University of Science and Technology (HUST), China. His research interests include computer vision, pattern recognition, video analytics, gesture and action analysis, large-scale visual search and mining. He received Best Paper Award from IEEE Trans. on Multimedia, Nanyang Assistant Professorship from NTU, and Outstanding EECS Ph.D. Thesis award from Northwestern University. He is currently Senior Area Editor of Journal of Visual Communications and Image Representation (JVCI), Associate Editor of IEEE Trans. on Image Processing (T-IP) and IEEE Trans. on Circuits and Systems for Video Technology (T-CSVT), and served as Guest Editor of International Journal of Computer Vision (IJCV). He is Program Co-Chair of IEEE Conf. on Multimedia Expo (ICME'18) and Steering Committee Member of ICME (2018-2019). He also served as Area Chair for CVPR, ICIP, ICPR, ACCV, ACM MM, WACV etc. He is a Fellow of International Association of Pattern Recognition (IAPR).
    
\end{IEEEbiography}





\end{document}